\documentclass[letterpaper, 10 pt, conference]{ieeeconf}  

\IEEEoverridecommandlockouts                              

\usepackage{cite}

\usepackage{amsmath,amssymb,amsfonts,amsthm}
\usepackage{algorithmic}
\usepackage{algorithm}
\usepackage{graphicx}
\usepackage{textcomp}
\usepackage{xcolor}
\usepackage{svg}
\usepackage{dsfont}
\usepackage{adjustbox}

\usepackage{xspace}
\usepackage{tikz}
\usepackage{hyperref}

\newcommand\copyrighttext{%
  \footnotesize \textcopyright 2025 IEEE. Personal use of this material is permitted.
  Permission from IEEE must be obtained for all other uses, in any current or future
  media, including reprinting/republishing this material for advertising or promotional
  purposes, creating new collective works, for resale or redistribution to servers or
  lists, or reuse of any copyrighted component of this work in other works.}
\newcommand\copyrightnotice{%
\begin{tikzpicture}[remember picture,overlay]
\node[anchor=south,yshift=10pt] at (current page.south) 
  {\fbox{\parbox{\dimexpr\textwidth-\fboxsep-\fboxrule\relax}{\copyrighttext}}};
\end{tikzpicture}%
}
\providecommand{\FullStop}{\text{~\@.\xspace}}
\providecommand{\Comma}{\text{~,\xspace}}

\newtheorem{proposition}{Proposition}
\newtheorem{lemma}{Lemma}

\title{\LARGE \bf
Efficient Collision Detection for Long and Slender Robotic Links in Euclidean Distance Fields: Application to a Forestry Crane
}

\author{Marc-Philip Ecker$^{1,2}$, Bernhard Bischof$^{2}$, Minh Nhat Vu$^{1,2}$, Christoph Fröhlich$^{2}$,\\ Tobias Glück$^{2}$ and Wolfgang Kemmetmüller$^{1}$
\thanks{$^{1}$Marc-Philip Ecker, Minh Nhat Vu and Wolfgang Kemmetmüller are with the
Automation \& Control Institute (ACIN), TU Wien, 1040 Vienna, Austria
        {\tt\small \{ecker,vu,kemmetmueller\}@acin.tuwien.ac.at}}%
\thanks{$^{2}$Marc-Philip Ecker, Bernhard Bischof, Minh Nhat Vu, Christoph Fröhlich and Tobias Glück are with the Center for Vision, Automation \& Control,
AIT Austrian Institute of Technology GmbH, 1210 Vienna, Austria
        {\tt\small \{marc-philip.ecker,bernhard.bischof,minh.vu,\newline christoph.froehlich,tobias.glueck\}@ait.ac.at}}%
}

\begin{document}
\maketitle
\copyrightnotice
\thispagestyle{empty}
\pagestyle{empty}

\begin{abstract}
Collision-free motion planning in complex outdoor environments relies heavily on perceiving the surroundings through exteroceptive sensors. A widely used approach represents the environment as a voxelized Euclidean distance field, where robots are typically approximated by spheres. However, for large-scale manipulators such as forestry cranes, which feature long and slender links, this conventional spherical approximation becomes inefficient and inaccurate.

This work presents a novel collision detection algorithm specifically designed to exploit the elongated structure of such manipulators, significantly enhancing the computational efficiency of motion planning algorithms. Unlike traditional sphere decomposition methods, our approach not only improves computational efficiency but also naturally eliminates the need to fine-tune the approximation accuracy as an additional parameter. We validate the algorithm’s effectiveness using real-world LiDAR data from a forestry crane application, as well as simulated environment data.
\end{abstract}
\begin{figure}[t]
\centering
\adjustbox{trim=2.342cm 2.4cm 2.342cm 0cm, clip}{
\includegraphics[scale=0.25]{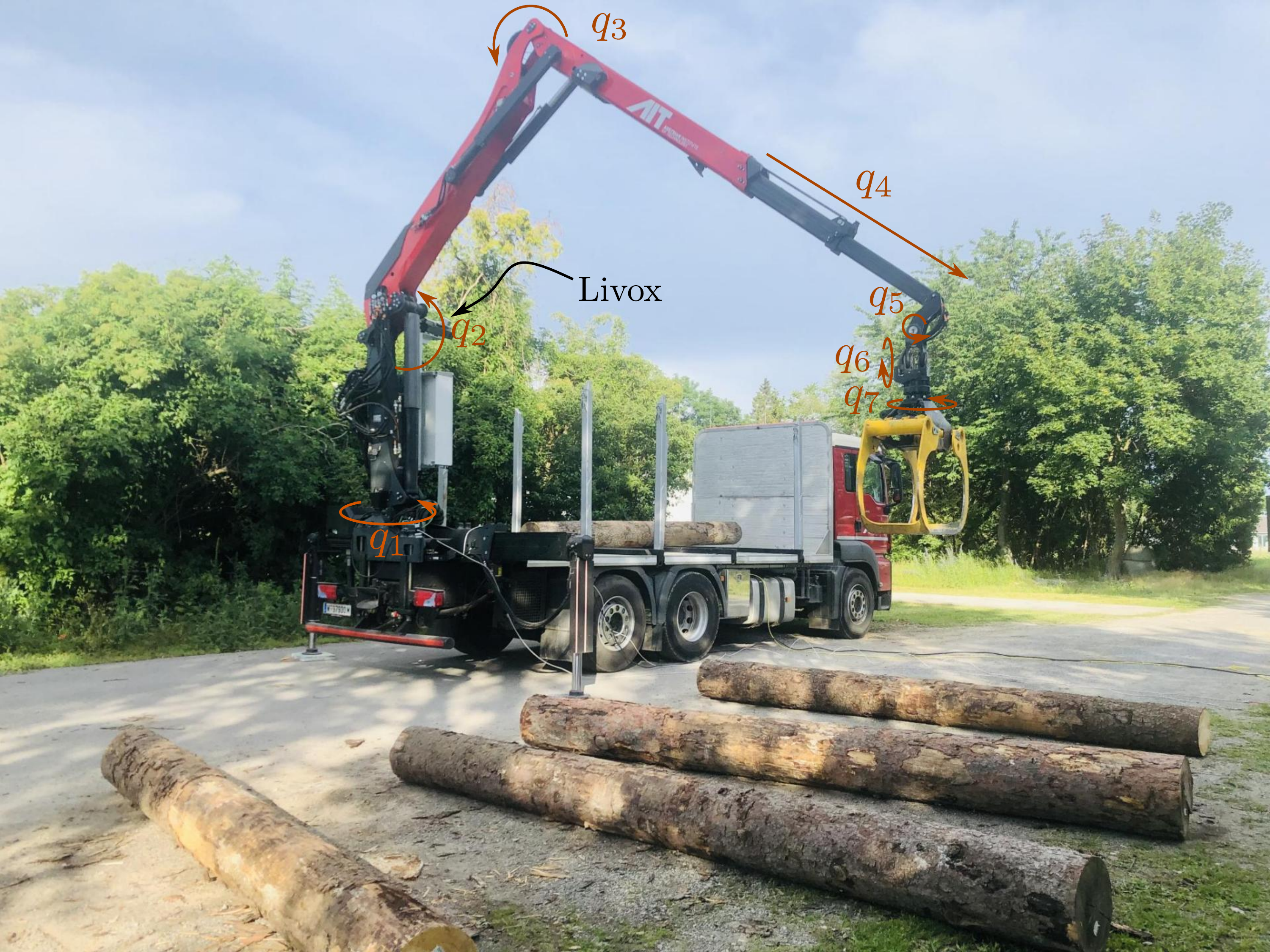}
}
\includegraphics[trim=4cm 0cm 8.2cm 0cm,clip,scale=0.25]{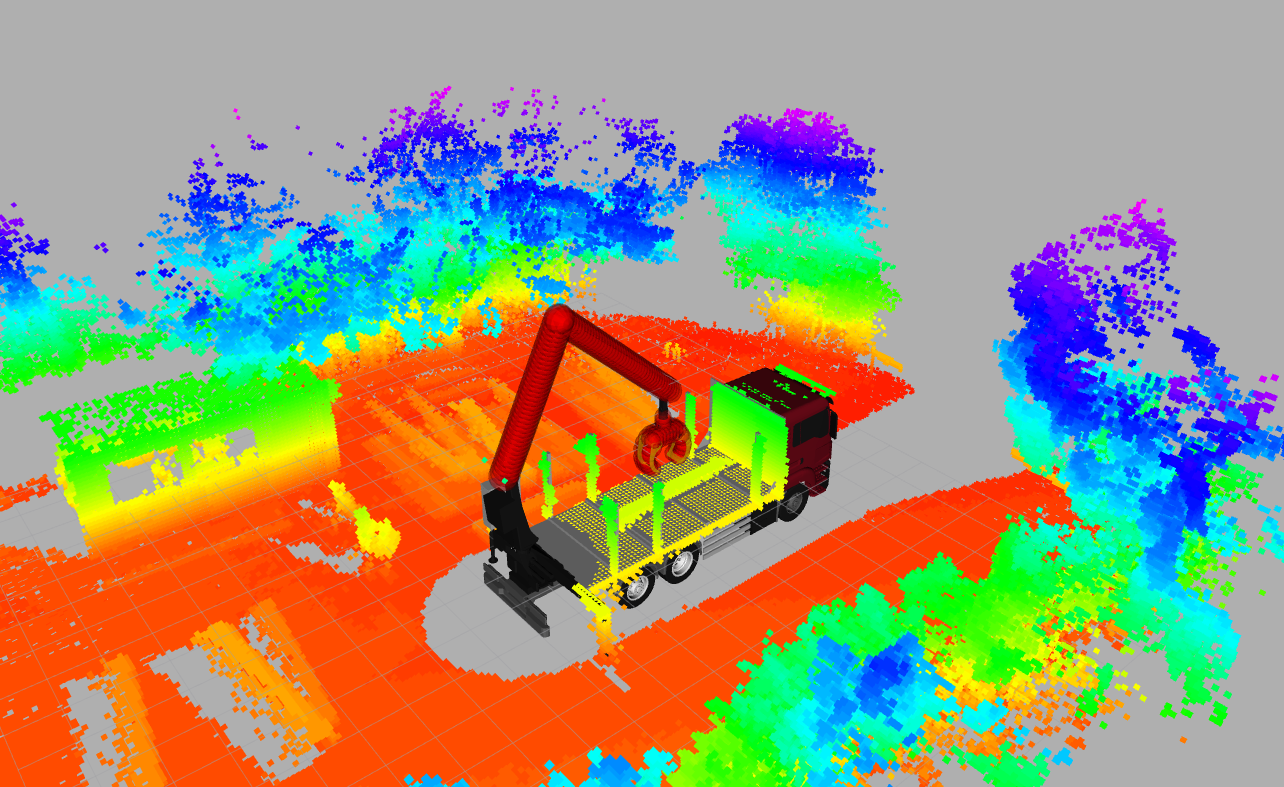}
\caption{Top: forestry crane considered for the evaluation. Bottom: occupied voxels of the environment map and collision spheres.}
\label{fig:KinematicChain}
\end{figure}

\section{Introduction}
Collision-free motion planning is a fundamental component of any autonomous robotic system. In controlled environments, obstacles can often be approximated as convex shapes, cf. \cite{schulman:2014}, allowing collision checks to be efficiently performed using algorithms such as the GJK algorithm \cite{gjk:1990,montaut:2024}. In contrast, large-scale manipulators, such as the forestry crane, illustrated in Fig.~\ref{fig:KinematicChain}, operate in complex outdoor environments that are not known in advance. In these scenarios, collision avoidance relies on environmental perception through exteroceptive sensors, such as Light Detection and Ranging (LiDAR) or cameras. The raw sensor data must then be processed into an environment representation suitable for collision-free motion planning.

Voxelized maps are a widely used solution for storing and processing environmental data. A notable approach is the Euclidean distance field (EDF), which encodes the shortest Euclidean distance to the nearest occupied point for each voxel. For instance, the lower portion of Fig.~\ref{fig:KinematicChain} visualizes the occupied voxels of an EDF constructed from LiDAR measurements of the forestry crane. In many mobile robotic applications, collision queries can be reduced to a simple lookup for a single point in the map \cite{oleynikova:2016}. However, for robots with extended structures, these geometries must be approximated by multiple spheres, cf.~\cite{ratliff:2009,chiu:2022,jelavic:2023}. This is especially true for large-scale manipulators, such as the forestry crane shown in Figure~\ref{fig:KinematicChain}, which consists of very long and slender links. Accurately representing such structures requires a large number of collision spheres, significantly increasing computational complexity.

This work builds on our previous work \cite{ecker:2025} and introduces a novel collision detection algorithm tailored for large-scale manipulators operating in Euclidean distance field representations. The proposed method is validated on real-world sensor data, demonstrating its efficiency in applications involving the forestry crane.

\subsection{Related Work}
Several efficient algorithms for collision detection between convex shapes exist, see, e.g., \cite{gjk:1990,montaut:2024,ericson:2005}. However, the environment has to be composed of convex shapes, which is typically not reasonable for complex environments perceived by exteroceptive sensor.
Contrary, Voxelized EDF maps are information rich representations, allowing online computation from point cloud data, see~\cite{oleynikova:2017,han:2019,pan:2022,millane:2024}. Each voxel in the map stores the distance to the closest surface boundary point. Hence, collision avoidance for extended structures requires a sphere decomposition of the robot, cf.~\cite{ratliff:2009,chiu:2022,jelavic:2023}. Covariant hamiltonian optimization (CHOMP) \cite{ratliff:2009} is an early approach that uses sphere approximations of a manipulator for distance computation in an EDF. In \cite{chiu:2022}, an automatic sphere decomposition algorithm is applied to achieve collision free model predictive control for whole body dynamic locomotion and manipulation.

Several works, such as \cite{wang:2006,voelz:2018}, tackle the problem of automatic sphere decomposition of given shapes. Others focus on efficient collision detection in sensor maps by leveraging SIMD parallelism \cite{ramsey:2024,thomason:2024}. 
In contrast to those strategies, our approach leverages the inherent structure of long links to develop an efficient and fast collision detection routine for large-scale manipulators. To the best of our knowledge, this specific approach has not yet been explored in the literature

\subsection{Contribution}
This paper introduces a novel collision detection routine to speed up motion planning for long and slender links in Euclidean distance fields. The computational efficiency of the approach is demonstrated in a forestry crane application, utilizing an EDF generated from LiDAR mounted on the crane. Furthermore, this work extends our previous work \cite{ecker:2025,ecker2:2025} by incorporating collision-free motion planning for a forestry crane in sensor-based maps.

\begin{figure}[t]
\centerline{\includegraphics[scale=0.85]{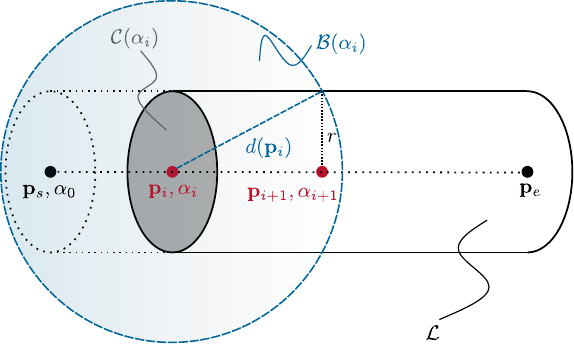}}
\caption{Uni-directional collision detection for long and slender links in EDF maps.}
\label{fig:CylindricalLink}
\end{figure}

\section{Collision Detection for Long Slender Links}
The long, slender links considered in this work are modeled as line segments with starting point \(\mathbf{p}_s\) and end point \(\mathbf{p}_e\). The core idea is to initiate collision detection from the starting point \(\mathbf{p}_s\). The Euclidean Distance Field (EDF) provides the shortest distance \(d(\mathbf{p}_s)\) from \(\mathbf{p}_s\) to the nearest obstacle in the environment. Consequently, as illustrated by the blue circle in Fig.~\ref{fig:CylindricalLink}, any point within a distance of \(d(\mathbf{p}_s)\) from \(\mathbf{p}_s\) must be free of collisions. Leveraging this property, we introduce two algorithms that iteratively determine a minimal set of query points for efficient collision detection.

\subsection{Formal Problem Statement}
For collision avoidance, the environment $\mathcal{E}\subseteq\mathbb{R}^3$ can be split in those points, that are occupied by obstacles $\mathcal{O}\subseteq\mathcal{E}$ and the ones that are free $\mathcal{F}\subseteq\mathcal{E}$, where $\mathcal{O}\cap\mathcal{F}=\emptyset$ and $\mathcal{F}\cup\mathcal{O}=\mathcal{E}$. The Eulidean signed distance for $\mathbf{p}\in\mathcal{F}$ is then defined as
\begin{align}
    d(\mathbf{p})=\inf_{\mathbf{p}_o\in\mathcal{O}}\|\mathbf{p}-\mathbf{p}_o\|\FullStop
\end{align}

We describe a link $\mathcal{L}$ of the manipulator in terms of its central axis
\begin{align}\label{eq:ParameterizedLine}
    \mathbf{l}(\alpha)=\mathbf{p}_s+\alpha\mathbf{t}\Comma\ \ \alpha\in[0,l]
\end{align}
with the start and end points $\mathbf{p}_s,\mathbf{p}_e\in\mathbb{R}^3$, the length $l=\|\mathbf{p}_e-\mathbf{p}_s\|$ and the direction $\mathbf{t}=\frac{\mathbf{p}_e-\mathbf{p}_s}{l}$. The link $\mathcal{L}$ is considered as collision-free, if a certain radius $r$ around the central axis $\mathbf{l}(\alpha)$ is collision-free, i.e.
\begin{align}\label{eq:CondCollisionFree}
\begin{aligned}
    d(\alpha)&>r\Comma & \forall\alpha\in[0,l]\Comma
\end{aligned}
\end{align}
where the abberiavated notation $d(\alpha)=d\big(\mathbf{l}(\alpha)\big)$ is used. Hence, the link is approximated by a capsule.

A circular cut, as illustrated in Figure~\ref{fig:CylindricalLink}, at $\alpha$ is defined as
\begin{align}\label{eq:CylindricalCut}
    \mathcal{C}(\alpha)=\{\mathbf{p}\in\mathbb{R}^3:(\mathbf{p}-\mathbf{l}(\alpha))^{\mathrm{T}}\mathbf{t}=0,\|\mathbf{p}-\mathbf{l}(\alpha)\|\leq r\}\FullStop
\end{align}
We say that $\mathcal{L}$ is collision free in $(\alpha_1,\alpha_2)$, if $\mathcal{C}(\alpha)\subseteq\mathcal{F}$, $\forall\alpha\in(\alpha_1,\alpha_2)$. This implies that the entire link $\mathcal{L}$ is collision-free, i.e., \eqref{eq:CondCollisionFree} is satisfied, if and only if $\mathcal{L}$ is collision-free in $(0,l)$, $d(0)>r$ and $d(l)>r$.


\subsection{Uni-Directional Search}\label{sec:OneDirectionalSearch}
Starting from \(\alpha_0 = 0\) with \(\mathbf{l}(\alpha_0) = \mathbf{p}_s\), our algorithm iteratively computes the next query point \(\alpha_{i+1} = \alpha_i + \Delta\alpha_i\) on the line, based on \(\alpha_i\) and the minimum distance \(d_i = d(\alpha_i)\). As illustrated by the blue circle in Fig.~\ref{fig:CylindricalLink}, the definition of the EDF allows us to infer a neighborhood around the query point that must be free of collisions.
\begin{lemma}\label{thm:CollisionFreeBall}
Given the distance $d(\alpha_i)$ at $\mathbf{l}(\alpha_i)$. Then all points $\mathbf{p}\in\mathcal{B}(\alpha_i)$ are collision-free, where
\begin{align}\label{eq:DefinitionBall}
    \mathcal{B}(\alpha_i)=\{\mathbf{p}\in\mathbb{R}^3:\|\mathbf{p}-\mathbf{l}(\alpha_i)\|<d(\alpha_i)\}\FullStop
\end{align}
\end{lemma}
\begin{proof}
Assume $\mathbf{p}\in\mathcal{O}$ for a given point $\mathbf{p}\in\mathcal{B}(\alpha_i)$. This yields $\|\mathbf{p}-\mathbf{l}(\alpha_i)\|<d(\alpha_i)=\inf_{\mathbf{p}_o\in\mathcal{O}}\|\mathbf{p}_o-\mathbf{l}(\alpha_i)\|$, which contradicts with $\mathbf{p}\in\mathcal{O}$.
\end{proof}

As we show in the following Proposition, we can use Lemma~\ref{thm:CollisionFreeBall} to guarantee that if the link has distance $d_i>r$ at $\mathbf{l}(\alpha_i)$ the link is collision-free in $(\alpha_i-\Delta\alpha_i,\alpha_i+\Delta\alpha_i)$ for a certain $\Delta\alpha_i$. This is used in the algorithm to skip collision queries for all points in this interval.

\begin{proposition}\label{thm:CollisionFreeSegment}
    Given the distance $d_i=d(\alpha_i)$ at $\mathbf{l}(\alpha_i)$ and let $d_i>r$. Further let 
    \begin{align}\label{eq:AlphaUpdateRule}
        \Delta\alpha_i=\sqrt{d_i^2-r^2}\FullStop
    \end{align}
    Then $\mathcal{L}$ is collision-free in $(\alpha_i-\Delta\alpha_i,\alpha_i+\Delta\alpha_i)$.
\end{proposition}
\begin{proof} Let $\mathbf{p}\in\mathcal{C}(\alpha_i+\Delta\alpha)$, $\Delta\alpha\in(-\Delta\alpha_i,\Delta\alpha_i)$. From the definition of $\mathbf{l}(\alpha)$ in \eqref{eq:ParameterizedLine}, we conclude $\mathbf{l}(\alpha_i)=\mathbf{l}(\alpha_i+\Delta\alpha_i)-\Delta\alpha_i\mathbf{t}$. This yields
    \begin{align}\label{eq:ProofCollisionFreeSegment}
    \begin{aligned}
        \|\mathbf{p}-\mathbf{l}(\alpha_i)\|&=\|\mathbf{p}-\mathbf{l}(\alpha_i+\Delta\alpha)+\Delta\alpha\mathbf{t}\|\\
        &=\sqrt{\|\mathbf{p}-\mathbf{l}(\alpha_i+\Delta\alpha)\|^2 + \Delta\alpha^2}\\
        &< \sqrt{r^2+\Delta\alpha_i^2}=d_i\Comma
    \end{aligned}
    \end{align}
    where the third equality results from $\|\mathbf{t}\|=1$ and $(\mathbf{p}-\mathbf{l}(\alpha_i+\Delta\alpha_i))^{\mathrm{T}}\mathbf{t}=0$, cf. \eqref{eq:CylindricalCut}. The inequality follows from $\|\mathbf{p}-\mathbf{l}(\alpha)\|\leq r$ and $\Delta\alpha^2<\Delta\alpha_i^2$. Hence, we conclude that $\mathbf{p}\in\mathcal{B}(\alpha_i)$ for all $\mathbf{p}\in\mathcal{C}(\alpha)$, $\alpha\in(\alpha_i-\Delta\alpha_i,\alpha_i+\Delta\alpha_i)$ and thus from Lemma~\ref{thm:CollisionFreeBall} we conclude that $\mathbf{p}$ is collision-free.
\end{proof}
Note that $[\alpha_i,\alpha_i+\Delta\alpha_i)\subset(\alpha_i-\Delta\alpha_i,\alpha_i+\Delta\alpha_i)$ and thus the link is collision-free in $[\alpha_i,\alpha_i+\Delta\alpha_i)$.

Proposition~\ref{thm:CollisionFreeSegment} allows to directly reason about intervals for $(\alpha_i-\Delta\alpha_i,\alpha_i+\Delta\alpha_i)$, that do not require additional collision queries. It is immediately clear that the nearest point that has to be queried lies at the boundary of this interval. Hence, $\alpha_{i+1}=\alpha_i+\Delta\alpha_i$ with $\Delta\alpha_i$ according to \eqref{eq:AlphaUpdateRule} is subsequently used.
\begin{algorithm}
\caption{Uni-Directional Search}\label{alg:OneDirectionalSearch}
\begin{algorithmic}[1]
\renewcommand{\algorithmicrequire}{\textbf{Input:}}
\renewcommand{\algorithmicensure}{\textbf{Output:}}
\REQUIRE Actual configuration $\mathbf{q}\in\mathbb{R}^n$
\ENSURE  true if collision, false otherwise.
\STATE Compute $\mathbf{p}_e,\mathbf{p}_s$ for link $\mathcal{L}$ based on $\mathbf{q}$.
\STATE $\alpha\gets 0$
\STATE $\mathbf{p}\gets\mathbf l(\alpha_0)$
\WHILE{$\alpha\leq l$}
    \IF{$d(\alpha)\leq r$}
        \RETURN true
    \ENDIF
    \STATE $\Delta\alpha \gets \sqrt{d(\alpha)^2-r^2}$
    \STATE $\alpha \gets \alpha + \Delta\alpha$
\ENDWHILE
\IF{$d(l)\leq r$}
    \RETURN true
\ENDIF
\RETURN false
\end{algorithmic} 
\end{algorithm}

Algorithm~\ref{alg:OneDirectionalSearch} summarizes the uni-directional search. Starting at $\alpha_0=0$, Proposition~\ref{thm:CollisionFreeSegment} guarantees, that the line is collision free in $[0,\Delta\alpha_0)$. We conclude that the next point that has to be queried in $\alpha_1=\alpha_0+\Delta\alpha_0$. If $d_1>r$, again Proposition~\ref{thm:CollisionFreeSegment} guarantees no collision in $[0,\alpha_1+\Delta\alpha_1)$ and so on. If $\alpha_i\geq l$, then $[0,l]$ is guaranteed to be collision-free. Finally, a query of $d(l)$ is required, since $\alpha_i<l$ might be the last queried point.

\subsection{Bi-Directional Search}

\begin{figure}[h]
\centerline{\includegraphics[scale=0.85]{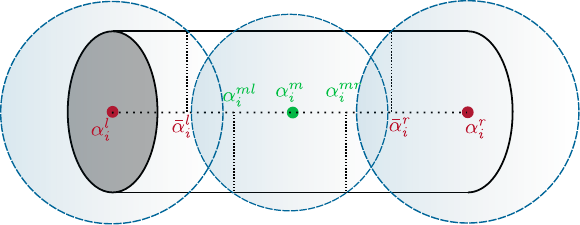}}
\caption{Bi-directional collision detection for long slender links in EDF maps.}
\label{fig:BiDirectionalCheck}
\end{figure}

One drawback of the uni-directional search presented in Section~\ref{sec:OneDirectionalSearch} is the significant overlap between the balls $\mathcal{B}(\alpha_i)$. To further reduce the number of collision queries, we introduce a bi-directional search strategy, as illustrated in Fig.~\ref{fig:BiDirectionalCheck}. The key idea is to systematically query points while minimizing overlap with previously queried points, thereby further reducing overlap of query points and thus the number of required collision checks.

The method maintains a queue $\mathcal{Q}$ of intervals that are not yet verified as collision free, initially set as $\mathcal{Q}=\{[0, l]\}$. At each iteration, we select an interval $[\alpha_i^l,\alpha_i^r]\in\mathcal{Q}$ and remove it. If either $d(\alpha_i^l)\leq r$ or $d(\alpha_i^r)\leq r$ the link is in collision and the algorithm stops. Otherwise, we compute $\Delta\alpha_i^l$ and $\Delta\alpha_i^r$ from \eqref{eq:AlphaUpdateRule} and update the points as $\bar{\alpha}_i^l=\alpha_i^l+\Delta\alpha_i^l$ and $\bar{\alpha}_i^r=\alpha_i^r-\Delta\alpha_i^r$ as shown in Fig.~\ref{fig:BiDirectionalCheck}.

From Proposition~\ref{thm:CollisionFreeSegment}, we can conclude that $[\alpha_i^l,\bar{\alpha}_i^l)$ and $(\bar{\alpha}_i^r,\alpha_i^r]$ are collision-free. Thus, if $\bar{\alpha}_i^l>\bar{\alpha}_i^r$ the whole interval $[\alpha_i^l,\alpha_i^r]$ is collision-free and nothing has to be added to the queue. Otherwise, we compute the midpoint $\alpha_i^m=\frac{1}{2}(\bar{\alpha}_i^l+\bar{\alpha}_i^r)$ and evaluate $d(\alpha_i^m)$ If $d(\alpha_i^m)>r$, Proposition~\ref{thm:CollisionFreeSegment} guarantees that the segment $(\alpha_i^{ml},\alpha_i^{mr})$, where $\alpha_i^{ml}=\alpha_i^m-\Delta\alpha_i^m$ and $\alpha_i^{mr}=\alpha_i^m+\Delta\alpha_i^m$, is collision-free. Again, we can conclude from $\bar{\alpha}_i^l>\alpha_i^{ml}$, that $[\alpha_i^l,\alpha_i^m]$ is collision-free. Otherwise we add $[\bar{\alpha}_i^l,\alpha_i^{ml}]$ to the queue, since it cannot be guaranteed as collision free. The same procedure is applied symmetrically to the right side. Algorithm~\ref{alg:BiDirectionalSearch} summarizes the bi-directional search strategy. 

\begin{algorithm}[t]
\caption{Bi-Directional Search}\label{alg:BiDirectionalSearch}
\begin{algorithmic}[1]
\renewcommand{\algorithmicrequire}{\textbf{Input:}}
\renewcommand{\algorithmicensure}{\textbf{Output:}}
\REQUIRE Actual configuration $\mathbf{q}\in\mathbb{R}^n$
\ENSURE  true if collision, false otherwise.
\STATE Compute $\mathbf{p}_e,\mathbf{p}_s$ for link $\mathcal{L}$ based on $\mathbf{q}$.
\STATE $\alpha^l\gets 0$, $\alpha^r\gets l$
\STATE $\mathcal{Q}\gets\{[\alpha^l,\alpha^r]\}$
\WHILE{$\mathcal{Q}\neq\emptyset$}
\STATE Choose and remove $(\alpha^l,\alpha^r)\in \mathcal{Q}$
\IF{$d(\alpha^l)\leq r\lor d(\alpha^r)\leq r$}
    \RETURN true
\ENDIF
\STATE $\Delta\alpha^l \gets \sqrt{d(\alpha^l)^2-r^2}$, $\Delta\alpha^r \gets 
\sqrt{d(\alpha^r)^2-r^2}$
\STATE $\bar{\alpha}^l\gets\alpha^l+\Delta\alpha^l$, $\bar{\alpha}^t\gets\alpha^r-\Delta\alpha^r$
\IF{$\bar{\alpha}^l < \bar{\alpha}^r$}
    \STATE $\alpha^m=\frac{1}{2}(\bar{\alpha}^l+\bar{\alpha}^r)$
    \IF{$d(\alpha^m)\leq r$}
        \RETURN true
    \ENDIF
    \STATE $\alpha^{ml}\gets\alpha^m-\sqrt{d(\alpha^m)^2-r^2}$
    \STATE $\alpha^{mr}\gets\alpha^m+\sqrt{d(\alpha^m)^2-r^2}$
    \IF {$\bar{\alpha}^l<\alpha^{ml}$}
        \STATE $\mathcal{Q}\gets\mathcal{Q}\cup\{[\bar{\alpha}^l,\alpha^{ml}]\}$
    \ENDIF
    \IF {$\alpha^{mr}<\bar{\alpha}^r$}
        \STATE $\mathcal{Q}\gets\mathcal{Q}\cup\{[\bar{\alpha}^{mr},\alpha^r]\}$
    \ENDIF
\ENDIF
\ENDWHILE
\RETURN false
\end{algorithmic} 
\end{algorithm}

\subsection{Consideration of Safety Distances}
Consider the case where we want to maintain a minimum safety distance \( d_s > 0 \) from the environment. This modifies the collision-free condition in \eqref{eq:CondCollisionFree} to
$d_i - d_s > r$.
Hence, there are two possible approaches: either subtracting the safety distance from the measured distance, i.e., \( \bar{d}_i = d_i - d_s \), or adding it to the link radius, \( \bar{r} = r + d_s \). Each approach results in a different update rule for \eqref{eq:AlphaUpdateRule}
\begin{subequations}\label{eq:SafetyTechniques}
\begin{align}
    \Delta\alpha_i^{(1)} &= \sqrt{\bar{d}_i^2 - r^2} = \sqrt{d_i^2 \textcolor{blue}{- 2d_i d_s + d_s^2} - r^2} \\
    \Delta\alpha_i^{(2)} &= \sqrt{d_i^2 - \bar{r}^2} = \sqrt{d_i^2 \textcolor{blue}{- 2r d_s - d_s^2} - r^2} \Comma
\end{align}  
\end{subequations}
where \( \Delta\alpha_i^{(1)} \) and \( \Delta\alpha_i^{(2)} \) are the update rules resulting from using \( \bar{d}_i \) and \( \bar{r} \), respectively. For clarity, the differences between these update rules are highlighted in blue.  

The key question is whether one variant is preferable. By evaluating the inequality \( \big(\Delta\alpha_i^{(1)}\big)^2 < \big(\Delta\alpha_i^{(2)}\big)^2 \), we obtain the condition
\begin{align}\label{eq:CondLargerUpdate}
d_i - d_s > r \ .
\end{align}

Hence, if condition \eqref{eq:CondLargerUpdate} holds, the update rule using \( \bar{r} \) results in larger updates. Since a violation of \eqref{eq:CondLargerUpdate} corresponds to a detected collision (where no update is needed), the approach using \( \bar{r} \) is preferable, as it generally leads to larger and more efficient update steps.

\begin{figure}
\centering
\includegraphics[trim=4cm 0cm 8.2cm 0cm,clip,scale=0.2]{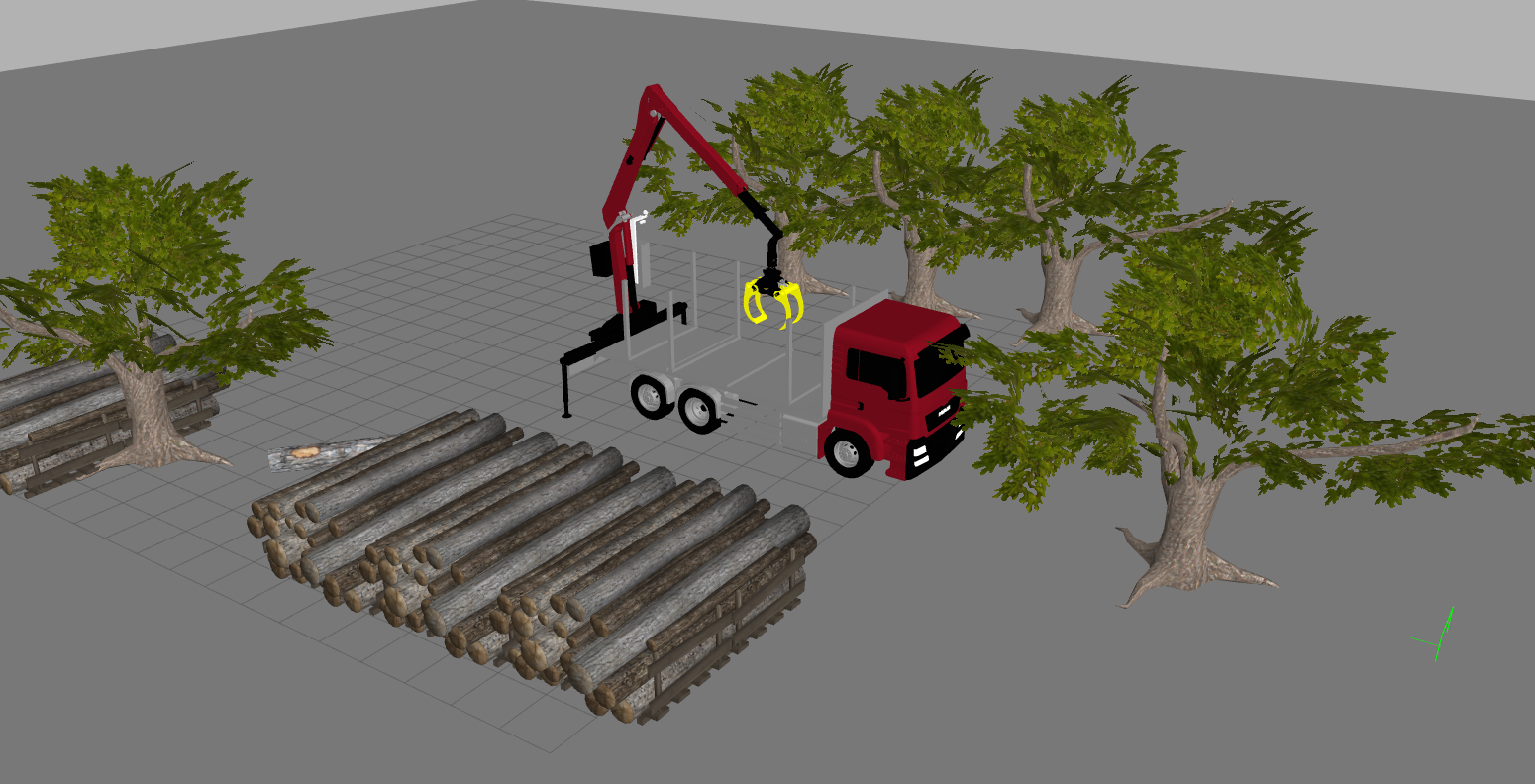}
\caption{Gazebo simulation environment.}
\label{fig:SimulationEnvironemnt}
\end{figure}

\begin{figure*}
\centering
\includegraphics[scale=0.26]{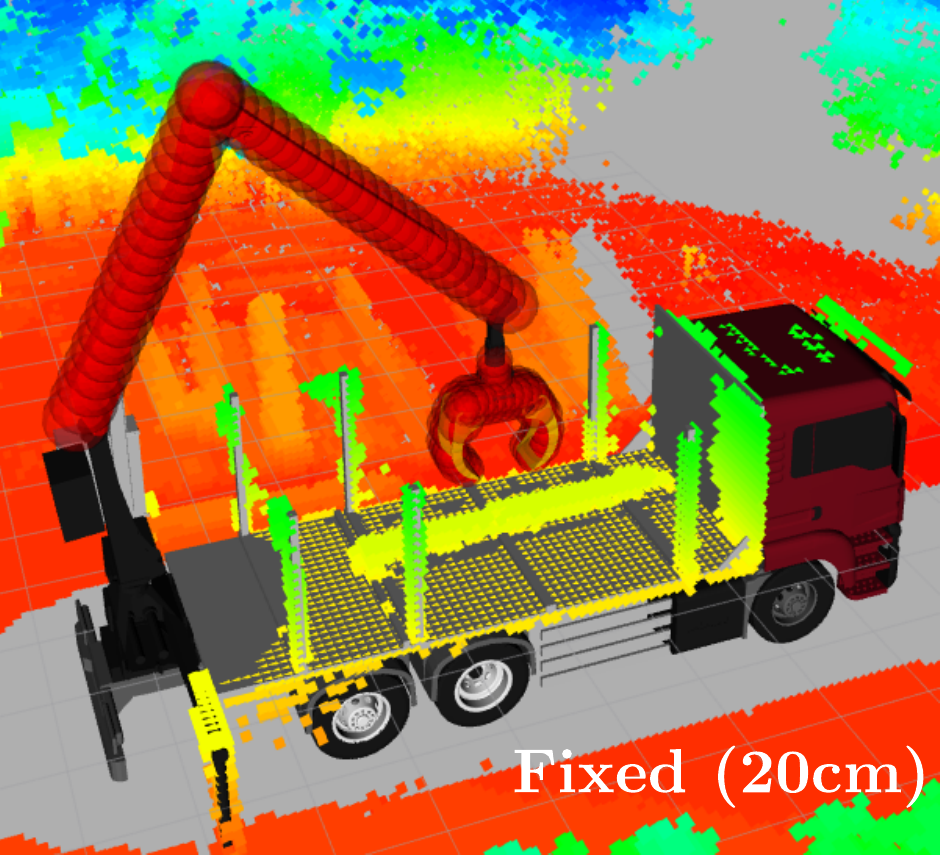}
\includegraphics[scale=0.26]{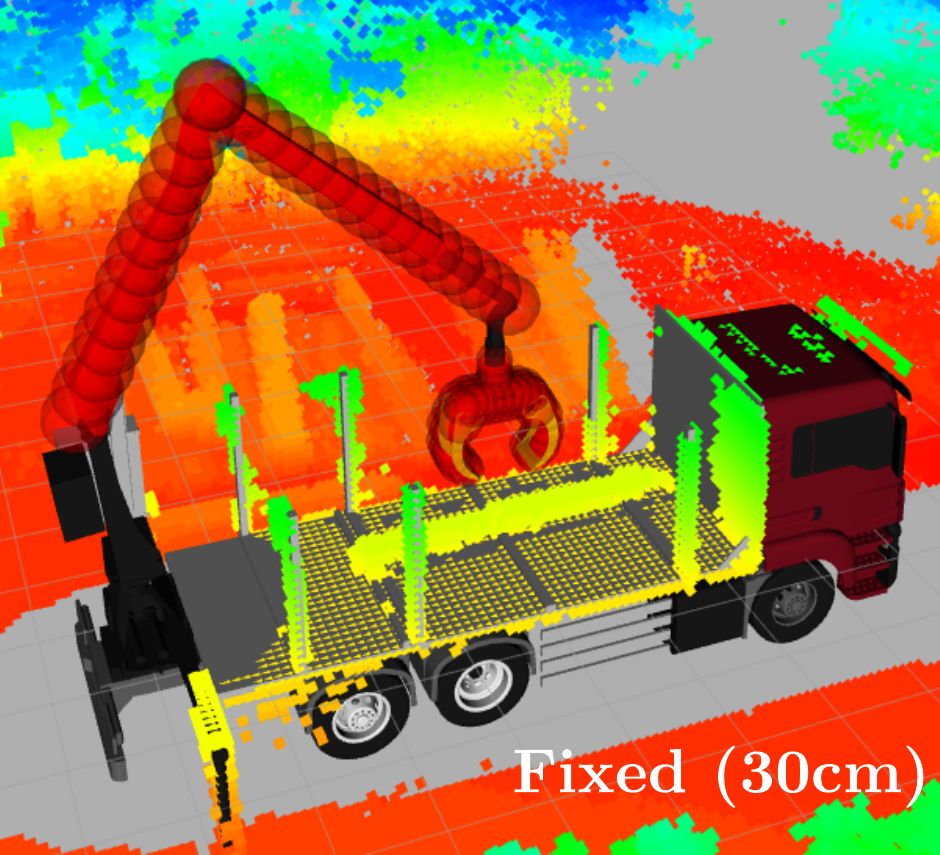}
\includegraphics[scale=0.26]{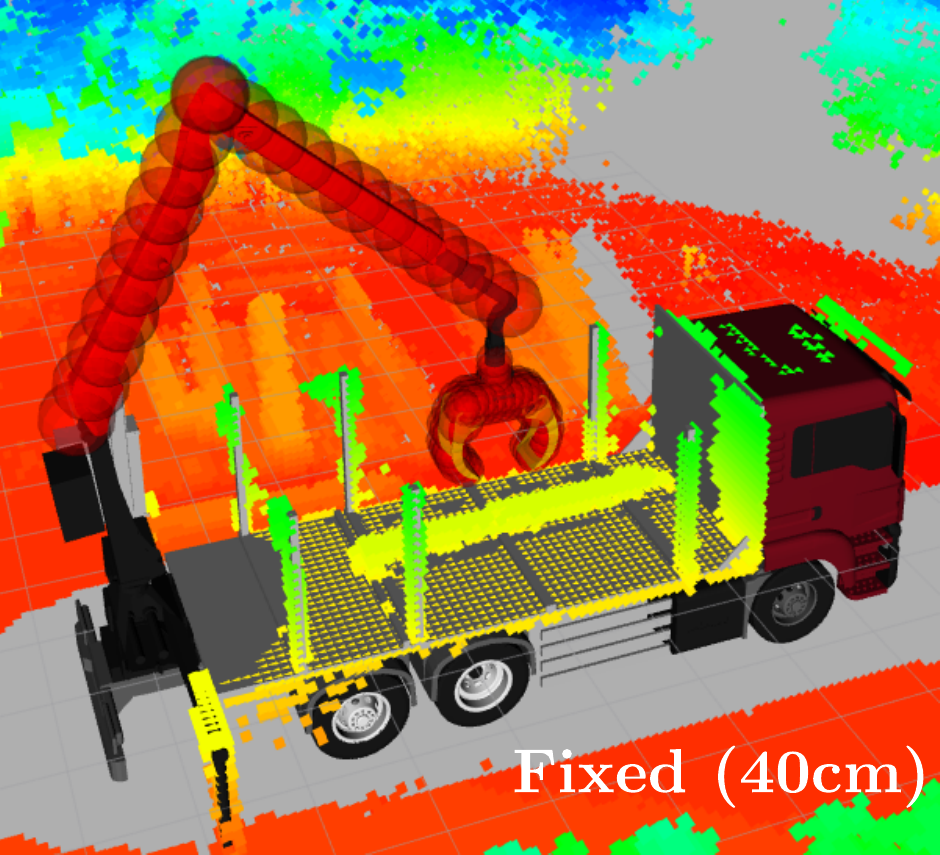}
\includegraphics[scale=0.26]{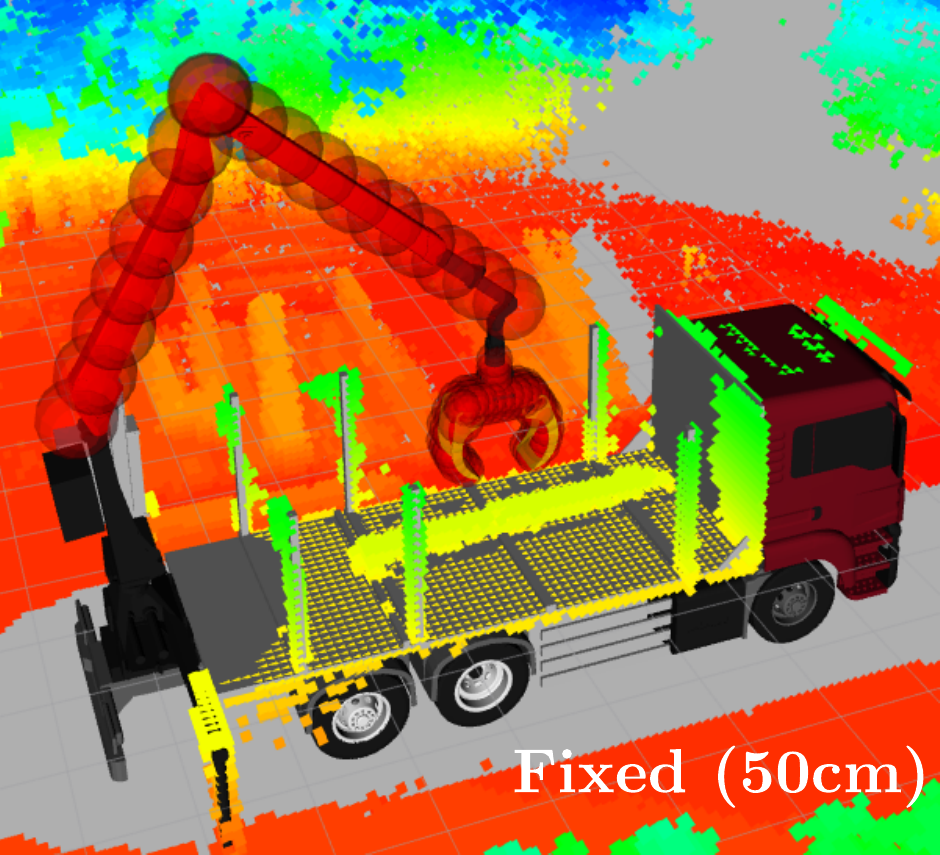}\\
\hfill\\
\includegraphics[scale=0.1975]{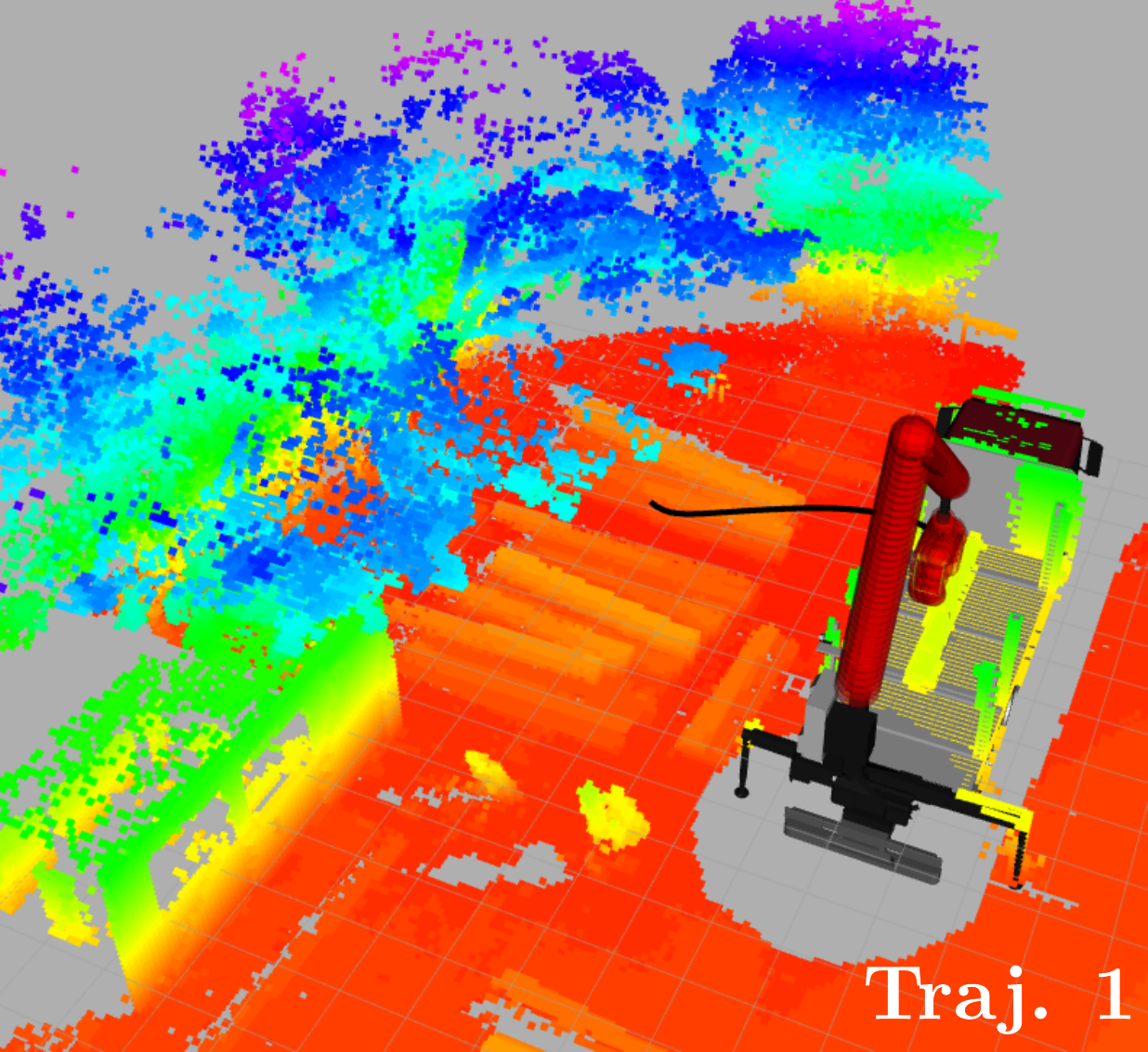}
\includegraphics[scale=0.1975]{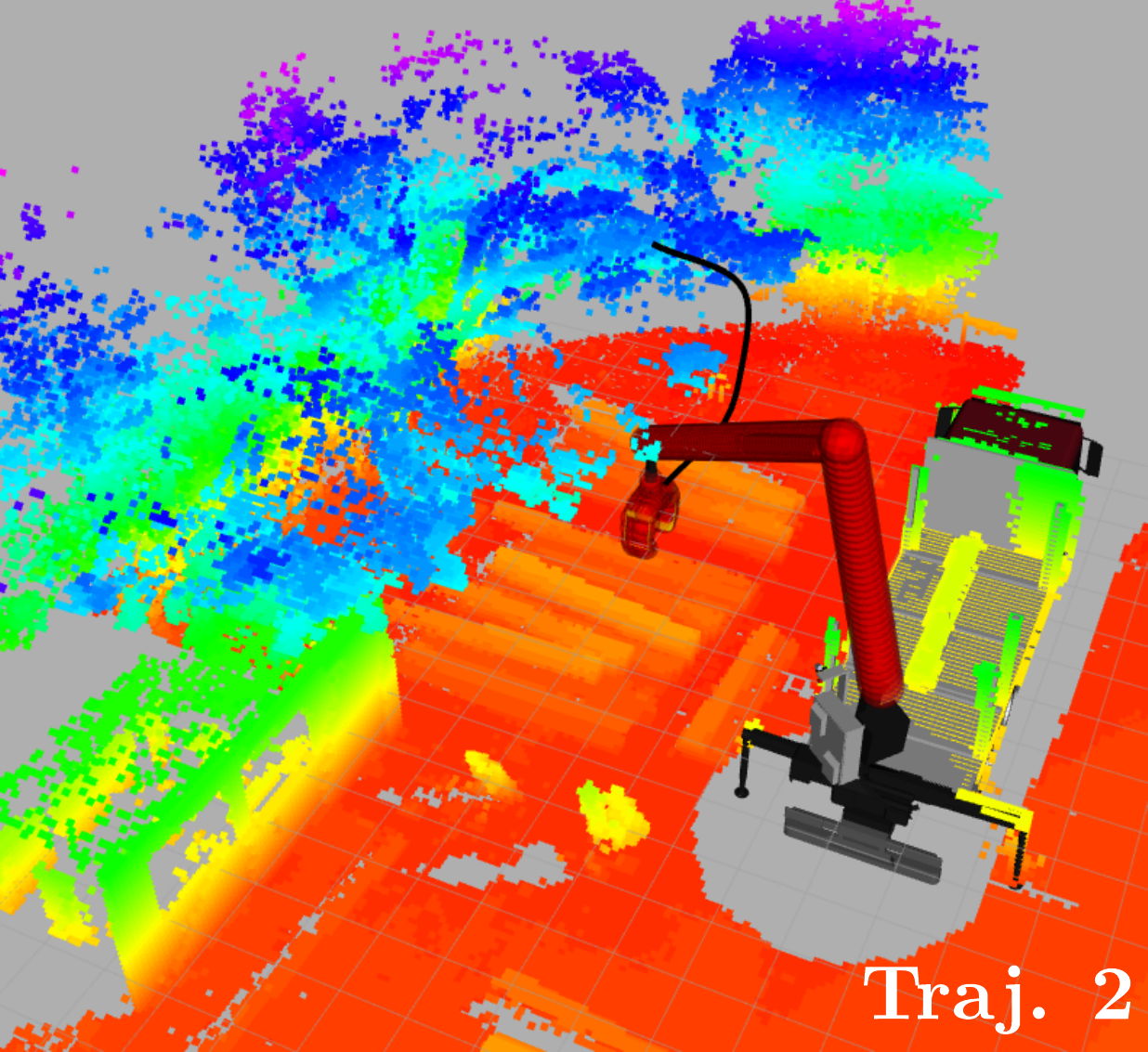}
\includegraphics[scale=0.1975]{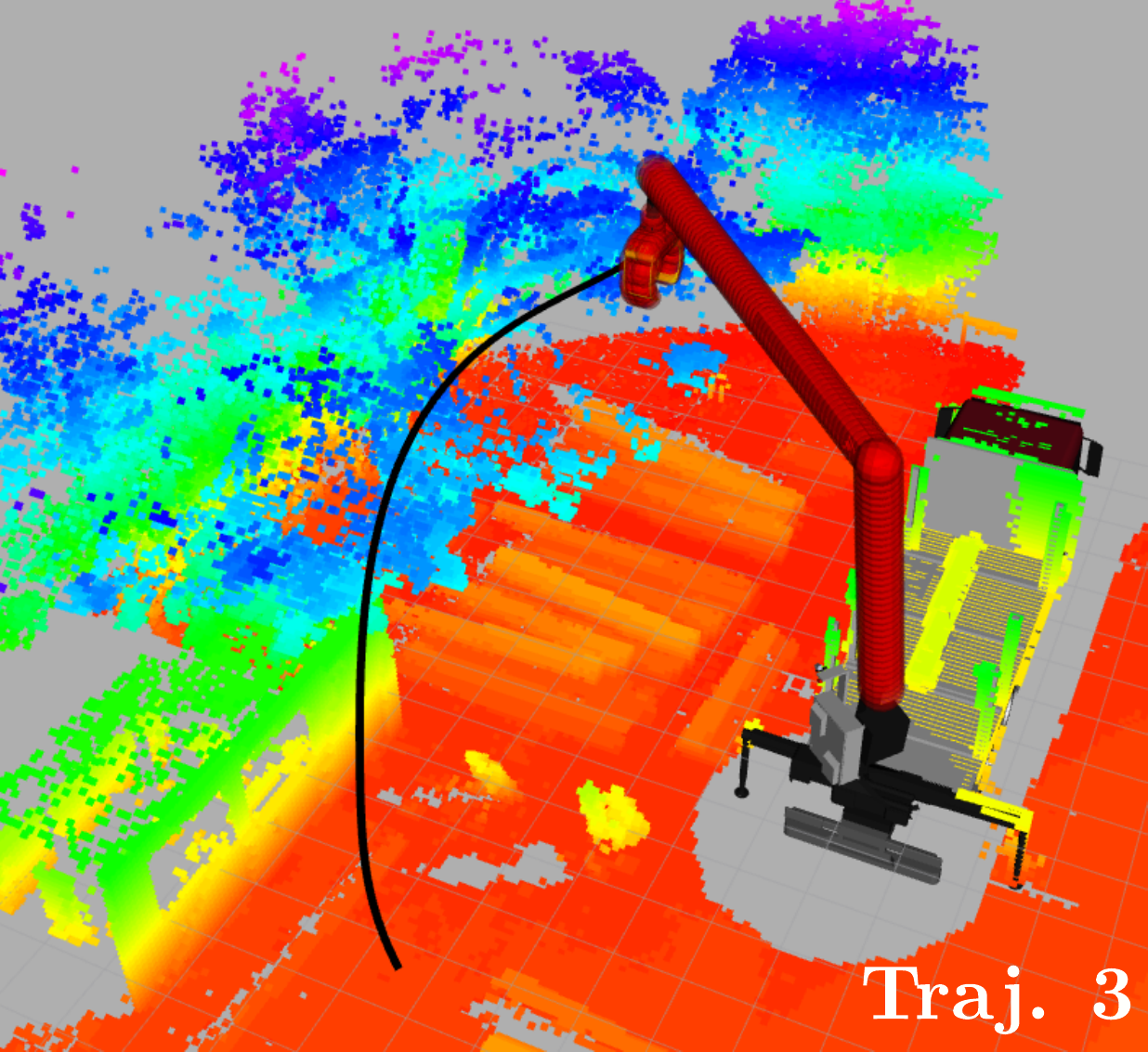}
\includegraphics[scale=0.1975]{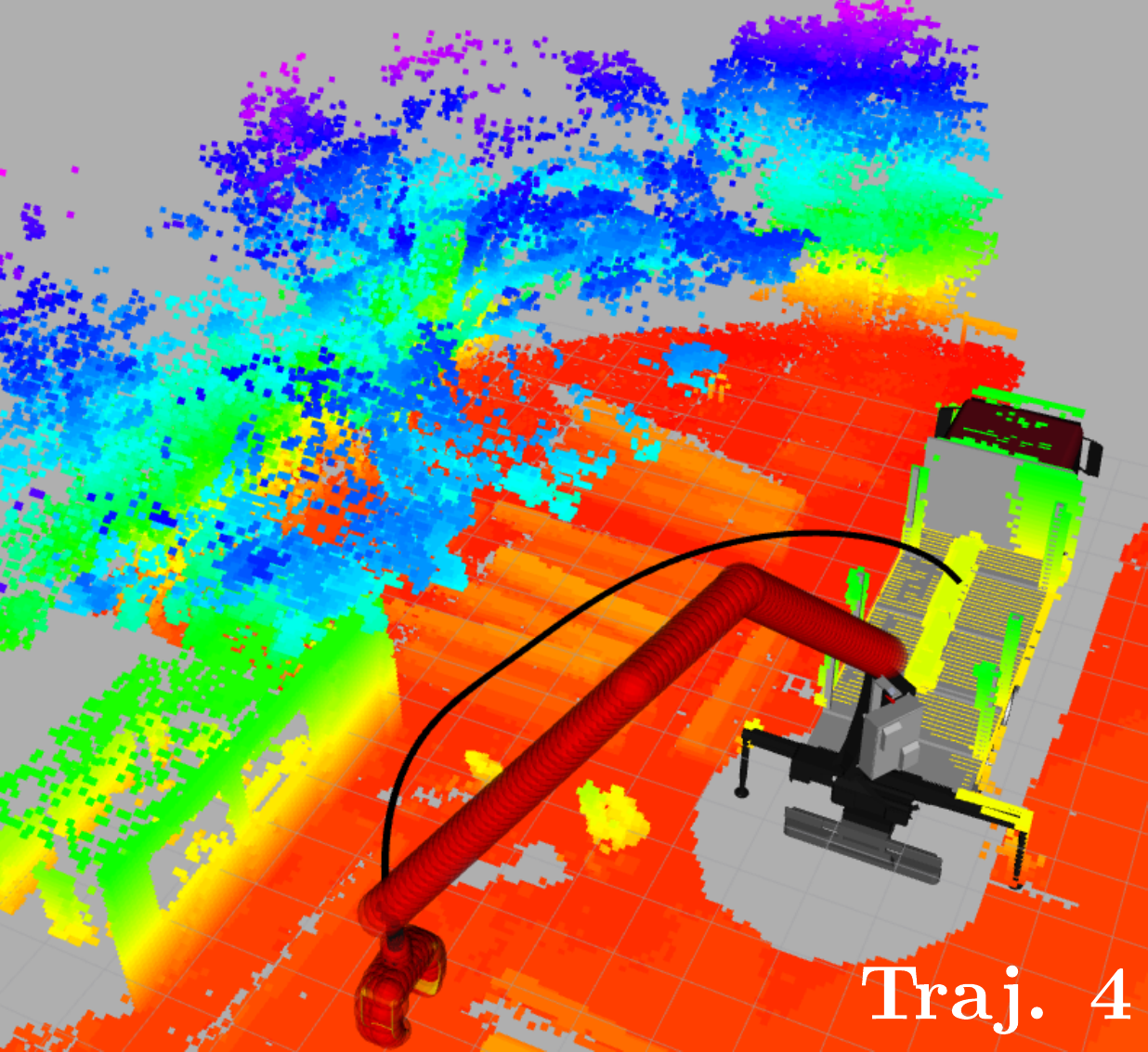}
\caption{Top: different sphere separation distances, 20cm - 50cm. Bottom: sample trajectories for the motion planning benchmark.}
\label{fig:SphereSeparationDistances}
\end{figure*}

\section{Results}
We evaluate our approach on the forestry crane depicted in Fig.~\ref{fig:KinematicChain}. All algorithms are implemented in 	\textsc{C++} using ROS 2, and all experiments are conducted on an Intel Core i7-10850H CPU @ 2.7 GHz with 12 cores.

\subsection{Experimental Setup}
The crane's generalized coordinates are denoted by $\mathbf{q}^{\mathrm{T}}=[q_1,\dots,q_7]$, as illustrated at the top of Fig.\ref{fig:KinematicChain}, where $q_5$ and $q_6$ represent passive joints. The crane consists of two long, slender link segments connecting $q_2$ to $q_3$ and $q_3$ to $q_5$, respectively. The first link maintains a fixed length of 3.5m, with a collision capsule radius of 0.32m, as visualized at the bottom of Fig.\ref{fig:KinematicChain}. In contrast, the second link's length depends on $q_4$, measuring 3.15m when $q_4=0$, and features a collision capsule radius of 0.3m.

The crane is equipped with a Livox Avia LiDAR \cite{livox:manual}, which is employed to compute an EDF in our test environment. Additionally, a simulated environment is incorporated, as depicted in Fig.~\ref{fig:SimulationEnvironemnt}. We utilize Octomap \cite{hornung:2013} to generate an occupancy map from both real and simulated sensor data. Subsequently, the voxel-based EDF map is computed using the algorithm proposed in \cite{han:2019}.
The crane model and motion planning algorithm is adopted from \cite{ecker:2025}.

\subsection{Monte Carlo Evaluation}
In the first experiments we sample $10^6$ random configurations and measure the runtimes as well as the number of collision queries for both our approaches. In addition to the real environment depicted in Fig.~\ref{fig:KinematicChain}, we include results from a Gazebo environment, cf. Fig.~\ref{fig:SimulationEnvironemnt}. For the fixed sphere decomposition, we benchmark different separation distances of the spheres, asserting that the intersection distance is equal to the link radius, cf. Fig.~\ref{fig:SphereSeparationDistances}. Note that the length of the second link depends on the value of the translational joint of the telescope arm. Our goal is to assess the performance of our collision detection routine in comparison to a fixed decomposition. Therefore, the crane's gripper is excluded from this experiment, as it cannot be accurately represented as a long link segment. Including it would distort the results, making it difficult to evaluate the true effectiveness of our approach. However, in Section~\ref{sec:MPBenchmark}, the gripper is included as part of a full motion planning problem.

\begin{table}[h]
    \centering
    \caption{Percentage of detected collisions.}\label{tab:NumCollisions}
    \begin{tabular}{||c||c|c||}
        \hline\hline
        \textbf{Method} & \multicolumn{2}{|c|}{\textbf{Collisions in $\%$}}\\
        \hline
        & Sim. & Real\\
        \hline\hline
        Bi-Directional & 30.21 & 34.12\\
        Uni-Directional & 30.21 & 34.12\\
        Fixed (10cm) & 30.34 & 34.24\\
        Fixed (20cm) & 30.67 & 34.65\\
        Fixed (30cm) & 31.21 & 35.77\\
        Fixed (40cm) & 31.92 & 36.54\\
        Fixed (50cm) & 32.78 & 41.22\\
        \hline\hline
    \end{tabular}
\end{table}
\subsubsection{Approximation accuracy} Table~\ref{tab:NumCollisions} presents the percentage of collisions observed in the sampled configuration data in both, the simulated and the real environment. As the separation distance between spheres decreases, the collision percentage for the fixed-sphere approximations gradually converges to that of the proposed methods. This occurs because the conservative approximation of the capsule using spheres necessitates an increase in sphere radius to ensure collision-free verification. In contrast, our approaches directly leverages the link structure, eliminating the need for such conservative approximations.

\subsubsection{Performance} Table~\ref{tab:StatisticsCollCheck} shows the corresponding mean runtimes and number of collision queries per configuration. It can be seen, that both of our algorithms outperform fixed sphere decomposition of the manipulator in all considered resolutions. Our bi-directional approach yields an runtime decrease of approximately 80\% compared to the 10cm resolution and 40\% compared to the 50cm resolution.
Additionally, we compare our approach to the Flexible Collision Library (FCL) \cite{pan:2012}, which utilizes an OctoMap representation of the environment. This method eliminates the need for sphere decomposition, allowing capsules to be directly used for collision detection. However, as shown in Table~\ref{tab:StatisticsCollCheck}, this results in a significant increase in computation time, as lookups in the Euclidean Distance Field are considerably more efficient.
\begin{table}[h]
    \centering
    \caption{Average Runtimes and number of collision queries.}\label{tab:StatisticsCollCheck}
    \begin{tabular}{||c||c|c||c|c||}
        \hline\hline
        \textbf{Method} & \multicolumn{2}{|c|}{\textbf{Runtime in $\mu s$}} & \multicolumn{2}{|c|}{\textbf{Collision Queries}}\\
        \hline
        & Sim. & Real & Sim. & Real\\
        \hline\hline
        Bi-Directional & \textbf{0.69} & \textbf{0.68} & \textbf{5.51} & \textbf{5.31}\\
        Uni-Directional & 0.74 & 0.72 & 7.11 & 7.05\\
        Fixed (10cm) & 3.56 & 3.81 & 84.68 & 86.13\\
        Fixed (20cm) & 2.13 & 2.37 & 44.07 & 44.82\\
        Fixed (30cm) & 1.58 & 1.64 & 30.17 & 30.69\\
        Fixed (40cm) & 1.32 & 1.36 & 23.20 & 23.62\\
        Fixed (50cm) & 1.12 & 1.22 & 18.81 & 19.16\\
        Octree (FCL) & 23.98 & 24.01 & - & -\\
        \hline\hline
    \end{tabular}
\end{table}

\subsubsection{Effect of link length}\label{sec:EffectLinkLength}
\begin{figure}[t]
\adjustbox{trim=0.7cm 0.3cm 0cm 0.55cm, clip}{\includegraphics[scale=0.65]{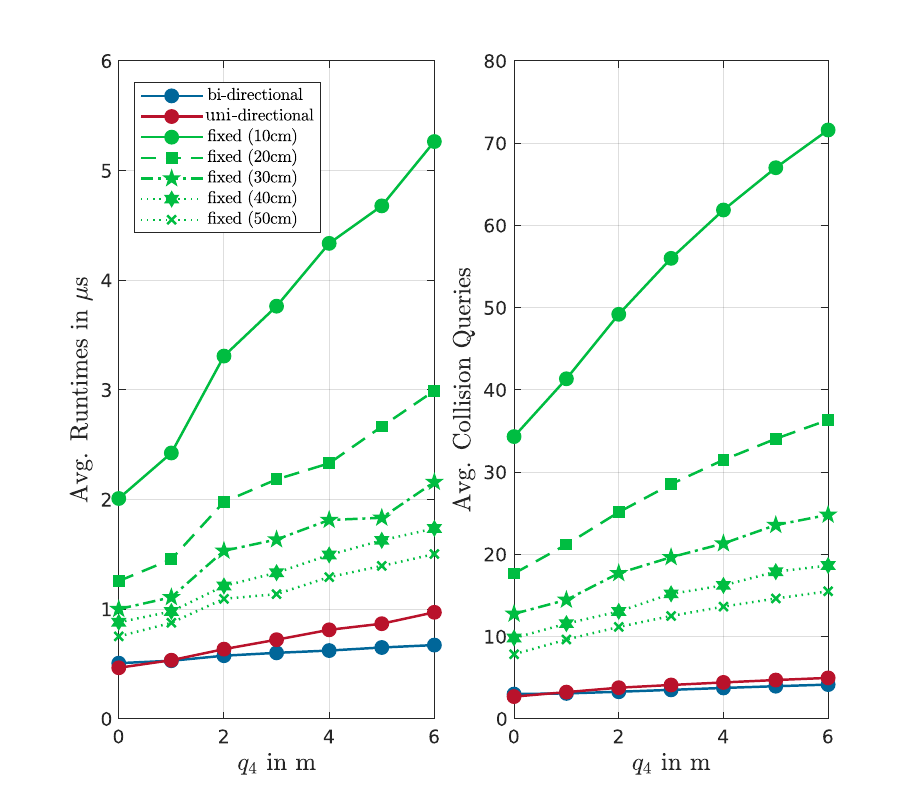}}
\caption{Average runtimes and number of collision queries for different link lengths.}
\label{fig:EffectLinkLength}
\end{figure}
To examine the impact of varying link lengths, we consider the telescope arm as a single link. The arm length \( q_4 \) is fixed at different values, and both the average collision detection runtime and the average number of collision queries per configuration are measured. The results are illustrated in Fig.~\ref{fig:EffectLinkLength}.  
All variants exhibit an approximately linear increase in runtime, with a slight decrease for longer links. This reduction is due to the higher frequency of collisions, which allow for early termination before the entire link is queried.
Among the evaluated methods, the bi-directional approach demonstrates the best scalability with increasing link length, outperforming both fixed-size decompositions and the uni-directional approach for longer links. However, for shorter links, the uni-directional approach offers a slight advantage.

\subsubsection{Effect of link radius and safety distance}\label{sec:EffectLinkRadius}
To assess the impact of varying link radii and different safety distance handling techniques \eqref{eq:SafetyTechniques}, we again consider the telescope arm as a single link with a fixed length of \( q_4 = 2 \)m.  
Fig.~\ref{fig:EffectRadii} illustrates the average runtime and the number of collision queries for different link radii, which are determined by the safety distances \( d_s \). Both metrics show an approximately linear decrease as the link radius increases. This decline occurs because larger radii lead to a higher frequency of collisions, enabling early termination before the entire link is queried. Furthermore, the reduction is more pronounced for fixed sphere decompositions with shorter separation distances.

\begin{figure}[h]
\adjustbox{trim=0.4cm 0.08cm 0cm 0.4cm, clip}{\includegraphics[scale=0.62]{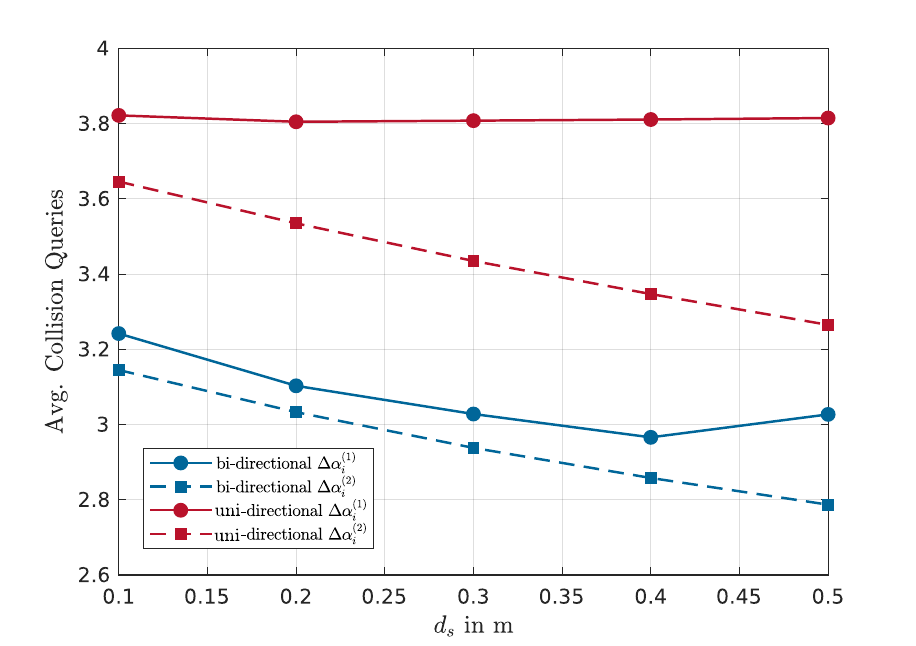}}
\caption{Comparison of the number of collision queries for different safety distance handling techniques \eqref{eq:SafetyTechniques}.}
\label{fig:EffectSafety}
\end{figure}

Fig.~\ref{fig:EffectSafety} compares the average number of collision queries for the two safety distance handling methods described in \eqref{eq:SafetyTechniques}. The results indicate that, on average, incorporating the safety distance into the link radius results in fewer distance queries and, consequently, larger update steps.

\subsubsection{Discussion}
\begin{figure}[t]
\adjustbox{trim=0.58cm 0.3cm 0cm 0.55cm, clip}{\includegraphics[scale=0.65]{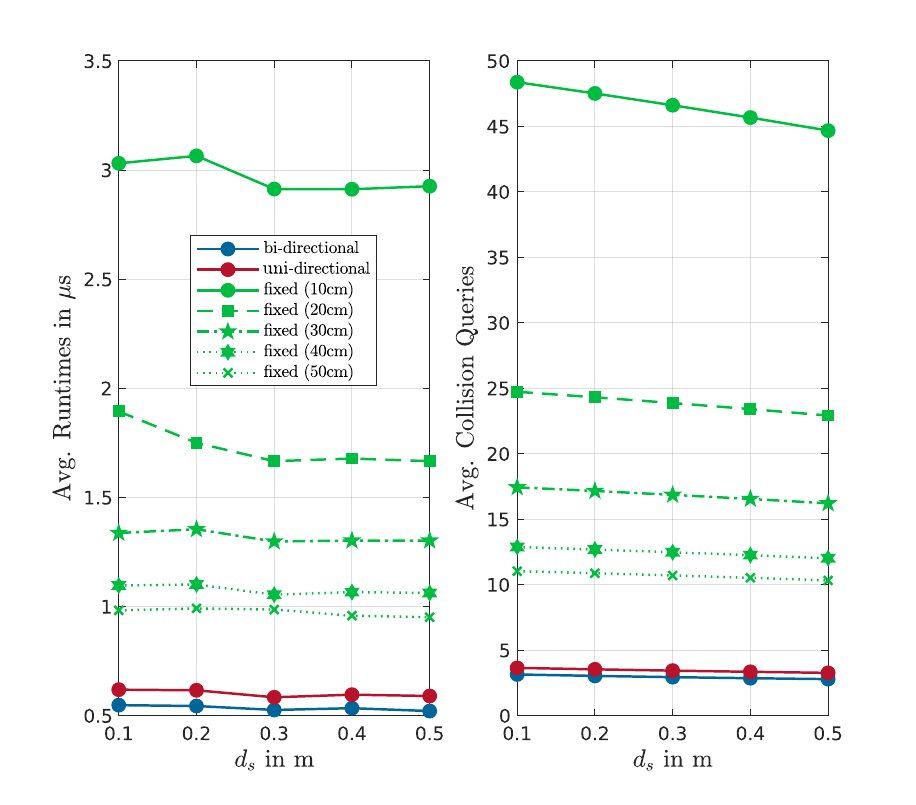}}
\caption{Average runtimes and number of collision queries for different link radii.}
\label{fig:EffectRadii}
\end{figure}

The results from Sections~\ref{sec:EffectLinkLength} and \ref{sec:EffectLinkRadius}, along with Fig.~\ref{fig:EffectLinkLength} and \ref{fig:EffectRadii}, indicate that the performance improvement of the proposed approaches is highly application-dependent. Specifically, it depends on factors such as link length and sphere resolution (Fig.~\ref{fig:EffectLinkLength}) as well as the link radius (as shown by the larger decrease rate in Fig.~\ref{fig:EffectRadii}). 
For the forestry crane application in our test environment, the average number of collision queries presented in Table~\ref{tab:StatisticsCollCheck} suggests that a fixed sphere decomposition with 5-8 spheres for both links would theoretically yield similar results as our approaches (disregarding additional computational overhead). However, this would yield a very conservative and impractical approximation of the crane’s geometry.
Additionally, as shown in Table~\ref{tab:NumCollisions} and Table~\ref{tab:StatisticsCollCheck}, determining the optimal sphere approximation accuracy in advance proves challenging. The algorithm designer must carefully balance accuracy and computational efficiency, as a coarse approximation reduces computation time but an overly conservative choice excessively restricts the crane's feasible motion areas. Our proposed solutions inherently eliminate the need to find this trade-off.

\begin{table*}[h]
    \centering
    \caption{Average runtimes in a Motion Planning application for the forestry crane.}\label{tab:MPBenchmark}
    \begin{tabular}{||c||c|c|c|c||c|c|c|c||}
        \hline\hline
        \textbf{Method} & \multicolumn{4}{|c|}{\textbf{Runtime in $\mathrm{s}$}}& \multicolumn{4}{|c|}{\textbf{Trajectory Duration in $\mathrm{s}$}}\\
        \hline
        & Traj. 1 & Traj. 2 & Traj. 3 & Traj. 4& Traj. 1 & Traj. 2 & Traj. 3 & Traj. 4\\
        \hline\hline
        Bi-Directional & \textbf{0.302} & \textbf{0.518} & \textbf{0.420} & \textbf{0.409} & 4.06 & 6.36 & 9.27 & 10.90 \\
        Uni-Directional & 0.320 & 0.522 & 0.435 & 0.429
            & 4.06 & 6.36 & 9.27 & 10.90\\
        Fixed (10cm) & 0.778 & 1.300 & 1.149 & 1.054 & 4.06 & 6.36 & 9.27 & 10.90\\
        Fixed (30cm) & 0.616 & 0.959 & 0.667 & 0.708 & 4.06 & 6.36 & 9.27 & 10.90\\\
        Fixed (50cm) & 0.423 & 0.728 & 0.534 & 0.587 & 4.06 & \textbf{6.47} & 9.27 & 10.90\\
        \hline\hline
    \end{tabular}
\end{table*}

\subsection{Motion Planning Benchmark}\label{sec:MPBenchmark}
Finally, we evaluate our approach on a motion planning problem for a forestry crane, based on the setup from \cite{ecker:2025}. Specifically, we use VP-STO \cite{janakowski:2023} to compute a global reference trajectory for the crane, utilizing 8 via-points, 100 evaluation points, and a sample size of 50. The algorithm is parallelized across 8 cores.

Table~\ref{tab:MPBenchmark} presents the average computation times of VP-STO with different collision detection techniques, based on 100 runs. For benchmarking, we selected four representative trajectories within the real sensor map, as shown in Fig.~\ref{fig:SphereSeparationDistances}. Similar results were obtained for other start and goal configurations, as well as in the simulated environment. Notably, the gripper is now incorporated into the model, as illustrated in Fig.~\ref{fig:KinematicChain}.

The bi-directional approach achieves a runtime reduction of approximately 60\% compared to the 10 cm sphere separation and 20\%–30\% compared to the 50 cm separation distance. However, in the 50 cm case, we observed that the planner struggled to find a solution for trajectory 2, as reflected in the increased trajectory durations in Table~\ref{tab:MPBenchmark} and a success rate drop to 89\%.

These results clearly highlight the advantages of integrating the proposed collision detection routine into a motion planning algorithm, significantly improving performance for real-world large-scale manipulator applications.

\section{Conclusion \& Future Work}
This work introduces efficient collision detection routines for long and slender links in Euclidean distance fields. We demonstrate the effectiveness and performance of the proposed solutions in a forestry crane application. By leveraging the inherent structure of long and thin links, our approach significantly reduces the number of collision queries, resulting in a performance improvement, depending on the sphere resolution,  of 40\%-80\% compared to a fixed sphere decomposition for collision detection for the timber crane. Furthermore, we benchmarked the methods in a motion planning scenario for a forestry crane, achieving a performance gain of 20\%-60\% in computation time, further validating their practical benefits.

While our motion planning algorithm already employs parallelization across multiple processors at the trajectory level, future work will focus on utilizing single instruction multiple data (SIMD) \cite{ramsey:2024} computation to enable parallelization at the configuration/link level, further enhancing efficiency.

\bibliographystyle{plain} 
\bibliography{refs} 

\end{document}